\def\year{2019}\relax
\documentclass[letterpaper]{article} 
\usepackage{aaai19}  
\usepackage{times}  
\usepackage{helvet}  
\usepackage{courier}  
\usepackage[hyphens]{url}  
\usepackage{graphicx}  
\usepackage{algorithm}
\usepackage{algorithmic}
\usepackage{amsthm}
\usepackage{amssymb}
\usepackage{multicol}
\usepackage{amsmath}

\usepackage[font=small]{caption}

\usepackage{color}
\usepackage[textsize=scriptsize]{todonotes}
\definecolor{blue}{RGB}{0, 93, 170}			
\definecolor{darkgreen}{RGB}{0, 102, 0}
\definecolor{orange}{RGB}{255, 160, 0} 

\newcommand{\ignore}[1]{}
\usepackage{tikz}
\usetikzlibrary{positioning,fit,arrows.meta,backgrounds}
\usetikzlibrary{shapes,decorations,arrows}
\usetikzlibrary{calc}

\tikzset{
    module/.style={%
        draw, rounded corners,
        minimum width=#1,
        minimum height=7mm,
        font={\footnotesize}
        },
    module/.default=2cm,
   	beats/.style={thick,->,>=stealth',draw}
}

\usepackage[font=small]{subcaption}
\usepackage[font=small]{caption}

\newcommand{\citet}[1]{\citeauthor{#1}~\shortcite{#1}}
\newcommand{\citep}{\cite}

\nocopyright
\title{Interpretable Multi-Objective Reinforcement Learning through Policy Orchestration}
\author{Ritesh Noothigattu$^\S$, Djallel Bouneffouf$^\dagger$, Nicholas Mattei$^\dagger$, Rachita Chandra$^\dagger$, \vspace{0.5em} \\
{\Large \bf Piyush Madan$^\dagger$, Kush Varshney$^\dagger$, Murray Campbell$^\dagger$, Moninder Singh$^\dagger$, Francesca Rossi$^\dagger$\thanks{On leave from the University of Padova.}}
\AND
{\rm $^\dagger$IBM Research} \\
IBM T.J. Watson Research Center \\
Yorktown Heights, NY, USA \\
\small{\{rachitac, krvarshn, mcam, moninder\}@us.ibm.com} \\ 
\small{\{djallel.bouneffouf, n.mattei, piyush.madan1, francesca.rossi2\}@ibm.com}
\And
{\rm $^\S$ Carnegie Mellon University}\\
Machine Learning Department\\
Pittsburgh, PA, USA \\
\small{riteshn@cmu.edu}
}

\begin{document}

\maketitle

\begin{abstract}
Autonomous cyber-physical agents and systems play an increasingly large role in our lives. To ensure that agents behave in ways aligned with the values of the societies in which they operate, we must develop techniques that allow these agents to not only maximize their reward in an environment, but also to learn and follow the implicit constraints of society.  These constraints and norms can come from any number of sources including regulations, business process guidelines, laws, ethical principles, social norms, and moral values.  
We detail a novel approach that uses inverse reinforcement learning to learn a set of unspecified constraints from demonstrations of the task, and reinforcement learning to learn to maximize the environment rewards.  More precisely, we assume that an agent can observe traces of behavior of members of the society but has no access to the explicit set of constraints that give rise to the observed behavior. Inverse reinforcement learning is used to learn such constraints, that are then combined with a possibly orthogonal value function through the use of a contextual bandit-based orchestrator that picks a contextually-appropriate choice between the two policies (constraint-based and environment reward-based) when taking actions.
The contextual bandit orchestrator allows the agent to mix policies in novel ways, taking the best actions from either a reward maximizing or constrained policy.  In addition, the orchestrator is transparent on  which policy is being employed at each time step.
We test our algorithms using a Pac-Man domain and show that the agent is able to learn to act optimally, act within the demonstrated constraints, and mix these two functions in complex ways. 
\end{abstract}

\section{Introduction}

Concerns about the ways in which cyber-physical and/or autonomous decision making systems behave when deployed in the real world are growing: what various stakeholder are worried about is that the systems achieves its goal in ways that are not considered acceptable according to values and norms of the impacted community, also called ``specification gaming'' behaviors. Thus, there is a growing need to understand how to constrain the actions of an AI system by providing boundaries within which the system must operate.

To tackle this problem, we may take inspiration from humans, who often constrain the decisions and actions they take according to a number of exogenous priorities, be they moral, ethical, religious, or business values \cite{Sen}, and we may want the systems we build to be restricted in their actions by similar principles \cite{conf/aaai/ArnoldKS17}. The overriding concern is that the autonomous agents we construct may not obey these values on their way to maximizing some objective function \cite{simonite2018}.

The idea of teaching machines right from wrong has become an important research topic in both AI \cite{yu2018building} and farther afield \cite{wallach2008moral}. Much of the research at the intersection of artificial intelligence and ethics falls under the heading of \emph{machine ethics}, i.e., adding ethics and/or constraints to a particular system's decision making process \cite{anderson2011machine}. One popular technique to handle these issues is called \emph{value alignment}, i.e., the idea that an agent can only pursue goals that follow values that are aligned to the human values and thus beneficial to humans \cite{russell2015research}.  

Another important notion for these autonomous decision making systems is the idea of \emph{transparency} or \emph{interpretability}, i.e., being able to see why the system made the choices it did.  \citet{theodorou2016my} observe that the  Engineering and Physical Science Research Council (EPSRC) Principles of Robotics dictates the implementation of transparency in robotic systems.  The authors go on to define transparency in a robotic or autonomous decision making system as, \emph{``... a mechanism to expose the decision making of the robot''}.

This still leaves open the question of how to provide the behavioral constraints to the agent.  A popular technique is called the \emph{bottom-up approach}, i.e., teaching a machine what is right and wrong by example \cite{allen2005artificial}. In this paper, we adopt this approach as we consider the case where only \emph{examples} of the correct behavior are available to the agent, and it must therefore learn from only these examples.  

We propose a framework which enables an agent to learn two policies: (1) $\pi_R$ which is a reward maximizing policy obtained through direct interaction with the world and (2) $\pi_C$ which is obtained via inverse reinforcement learning over demonstrations by humans or other agents of how to obey a set of behavioral constraints in the domain.  Our agent then uses a contextual-bandit-based orchestrator to learn to blend the policies in a way that maximizes a convex combination of the rewards and constraints. Within the RL community this can be seen as a particular type of apprenticeship learning \cite{abbeel2004apprenticeship} where the agent is learning how to be \emph{safe}, rather than only maximizing reward \cite{leike2017ai}.

One may argue that we should employ $\pi_C$ for all decisions as it will be more ``safe'' than employing $\pi_R$.  Indeed, although one could only use $\pi_C$ for the agent, there are a number of reasons to employ the orchestrator.  First, the humans or other demonstrators, may be good at demonstrating what not to do in a domain but may not provide examples of how best to maximize reward.  Second, the demonstrators may not be as creative as the agent when mixing the two policies \cite{VenturaG2018}. By allowing the orchestrator to learn when to apply which policy, the agent may be able to devise better ways to blend the policies, leading to behavior which both follows the constraints and achieves higher reward than any of the human demonstrations.  Third, we may not want to obtain demonstrations of what to do in all parts of the domain e.g., there may be dangerous or hard-to-model regions, or there may be mundane parts of the domain in which human demonstrations are too costly to obtain. In this case, having the agent learn through RL what to do in the non-demonstrated parts is of value. Finally, as we have argued, interpretability is an important feature of our system.  Although the policies themselves may not be directly interpretable (though there is recent work in this area \cite{VermaMSKC18,Guiliang05887}), our system does capture the notion of transparency and interpretability as we can see which policy is being applied in real time.

\smallskip
\noindent
\textbf{Contributions.\;} We propose and test a novel approach to teach machines to act in ways that achieve and compromise multiple objectives in a given environment. One objective is the desired goal and the other one is a set of behavioral constraints, learnt from examples.  Our technique uses aspects of both traditional reinforcement learning and inverse reinforcement learning to identify policies that both maximize rewards and follow particular constraints within an environment.  Our agent then blends these policies in novel and interpretable ways using an orchestrator based on the contextual bandits framework. We demonstrate the effectiveness of these techniques on the Pac-Man domain where the agent is able to learn both a reward maximizing and a constrained policy, and select between these policies in a transparent way based on context, to employ a policy that achieves high reward \emph{and} obeys the demonstrated constraints.

\section{Related Work}

Ensuring that our autonomous systems act in line with our values while achieving their objectives is a major research topic in AI.  These topics have gained popularity among a broad community including philosophers \cite{wallach2008moral} and non-profits \cite{russell2015research}. \citet{yu2018building} provide an overview of much of the recent research at major AI conferences on ethics in artificial intelligence. 

Agents may need to balance objectives and feedback from multiple sources when making decisions. One prominent example is the case of autonomous cars. There is extensive research from multidisciplinary groups into the questions of when autonomous cars should make lethal decisions \cite{bonnefon2016social}, how to aggregate societal preferences to make these decisions \cite{noothigattu2017voting}, and how to measure distances between these notions \cite{LoMaRoVe18,LoMaRoVe18a}. In a recommender systems setting, a parent or guardian may want the agent to not recommend certain types of movies to children, even if this recommendation could lead to a high reward \cite{BaBoMaRo18,balakrishnan2018incorporating}. Recently, as a compliment to their concrete problems in AI saftey which includes reward hacking and unintended side effects \cite{amodei2016concrete}, a DeepMind study has compiled a list of specification gaming examples, where very different agents game the given specification by behaving in unexpected (and undesired) ways.\footnote{38 AI ``specification gaming'' examples are available at: {\tiny \url{https://docs.google.com/spreadsheets/d/e/2PACX-1vRPiprOaC3HsCf5Tuum8bRfzYUiKLRqJmbOoC-32JorNdfyTiRRsR7Ea5eWtvsWzuxo8bjOxCG84dAg/pubhtml}}}

Within the field of reinforcement learning there has been specific work on ethical and interpretable RL. \citet{wu2017low} detail a system that is able to augment an existing RL system to behave ethically.  In their framework, the assumption is that, given a set of examples, most of the examples follow ethical guidelines.  The system updates the overall policy to obey the ethical guidelines learned from demonstrations using IRL.  However, in this system only one policy is maintained so it has no transparency. \citet{laroche2017reinforcement} introduce a system that is capable of selecting among a set of RL policies depending on context. They demonstrate an orchestrator that, given a set of policies for a particular domain, is able to assign a policy to control the next episode. However, this approach use the classical multi-armed bandit, so the state context is not considered on the choice of the policy.

Interpretable RL has received significant attention in recent years. \citet{LussP2016} introduce action constraints over states to enhance the interpretability of policies.  \citet{VermaMSKC18} present a reinforcement learning framework, called Programmatically Interpretable Reinforcement Learning (PIRL), that is designed to generate interpretable and verifiable agent policies. PIRL represents policies using a high-level, domain-specific programming language. Such programmatic policies have the benefit of being more easily interpreted than neural networks, and being amenable to verification by symbolic methods.  Additionally, \citet{Guiliang05887} introduce Linear Model U-trees to approximate neural network predictions. An LMUT is learned using a novel on-line algorithm that is well-suited for an active play setting, where the mimic learner observes an ongoing interaction between the neural net and the environment. Empirical  evaluation shows that an LMUT mimics a Q function substantially better than five baseline methods. The transparent tree structure of an LMUT facilitates understanding the  learned knowledge by analyzing feature influence, extracting rules, and highlighting the super-pixels in image inputs.

\section{Background}
\subsection{Reinforcement Learning}	\label{subsec:background-rl}
Reinforcement learning defines a class of algorithms solving problems modeled as a Markov decision process (MDP) \cite{Sutton1998}.

A Markov decision problem is usually denoted by the tuple $(\mathcal{S}, \mathcal{A}, \mathcal{T}, \mathcal{R}, \gamma)$, where 
\begin{itemize}
    \item $\mathcal{S}$ is a set of possible states
    \item $\mathcal{A}$ is a set of actions
    \item $\mathcal{T}$ is a transition function defined by $\mathcal{T}(s, a, s')=\Pr(s'\vert s,a)$, where $s, s'\in \mathcal{S}$ and $a\in \mathcal{A}$
    \item $\mathcal{R}: \mathcal{S}\times \mathcal{A} \times \mathcal{S}\mapsto \mathbb{R}$ is a reward function
    \item $\gamma$ is a discount factor that specifies how much long term reward is kept.
\end{itemize}

The goal in an MDP is to maximize the discounted long term reward received. Usually the infinite-horizon objective is considered:
\begin{align}
\max \sum_{t=0}^{\infty}\gamma^t \mathcal{R}(s_t,a_t, s_{t+1}).
\end{align}

Solutions come in the form of policies $\pi: \mathcal{S} \mapsto \mathcal{A}$, which specify what action the agent should take in any given state deterministically or stochastically. One way to solve this problem is through Q-learning with function approximation~\cite{bertsekas1996neuro}. The Q-value of a state-action pair, $\mathcal{Q}(s,a)$, is the expected future discounted reward for taking action $a \in \mathcal{A}$ in state $s \in \mathcal{S}$. A common method to handle very large state spaces is to approximate the $\mathcal{Q}$ function as a linear function of some features. Let $\boldsymbol{\psi}(s,a)$ denote relevant features of the state-action pair $\langle s, a \rangle$. Then, we assume $\mathcal{Q}(s,a) = \boldsymbol{\theta} \cdot \boldsymbol{\psi}(s,a)$, where $\boldsymbol{\theta}$ is an unknown vector to be learned by interacting with the environment. Every time the reinforcement learning agent takes action $a$ from state $s$, obtains immediate reward $r$ and reaches new state $s'$, the parameter $\boldsymbol{\theta}$ is updated using
\begin{equation} \label{eqn:q-update}
\begin{aligned}
\text{difference} &= \left[r + \gamma \max_{a'} \mathcal{Q}(s',a')\right] - \mathcal{Q}(s,a)\\
\theta_i &\gets \theta_i + \alpha \cdot \text{difference} \cdot \psi_i(s,a),
\end{aligned}
\end{equation}
where $\alpha$ is the learning rate. $\epsilon$-greedy is a common strategy used for exploration. That is, during the training phase, a random action is played with a probability of $\epsilon$ and the action with maximum Q-value is played otherwise. The agent follows this strategy and updates the parameter $\boldsymbol{\theta}$ according to Equation~\eqref{eqn:q-update} until the Q-value converge or for a large number of time-steps.

\subsection{Inverse Reinforcement Learning}	\label{subsec:background-irl}
IRL seeks to find the most likely reward function $\mathcal{R}_E$, which an expert $E$ is executing \cite{abbeel2004apprenticeship,Ng2000}. The IRL methods assume the presence of an expert that solves an MDP, where the MDP is fully known and observable by the learner except for the reward function. Since the state and action of the expert is fully observable by the learner, it has access to trajectories executed by the expert. A trajectory consists of a sequence of state and action pairs, $Tr =(s_0, a_0, s_1, a_1, \ldots, s_{L-1}, a_{L-1}, s_L)$, where $s_t$ is the state of the environment at time $t$, $a_t$ is the action played by the expert at the corresponding time and $L$ is the length of this trajectory. The learner is given access to $m$ such trajectories $\{Tr^{(1)}, Tr^{(2)}, \ldots, Tr^{(m)}\}$ to learn the reward function.
Since the space of all possible reward functions is extremely large, it is common to represent the reward function as a linear combination of $\ell > 0$ features. $\widehat{\mathcal{R}}_{\boldsymbol{w}}(s,a,s') = \sum_{i=1}^{\ell} w_i \phi_i(s,a,s')$, where $w_i$ are weights to be learned, and $\phi_i(s,a,s') \rightarrow \mathbb{R}$ is a feature function that maps a state-action-state tuple to a real value, denoting the value of a specific feature of this tuple~\cite{abbeel2004apprenticeship}. Current state-of-the-art IRL algorithms utilize feature expectations as a way of evaluating the quality of the learned reward function~\cite{abbeel2004apprenticeship}.
For a policy $\pi$, the feature expectations starting from state $s_o$ is defined as
\begin{align*}
\mu(\pi) = \mathbb{E}\left[\sum_{t=0}^\infty \gamma^t \boldsymbol{\phi}(s_t, a_t, s_{t+1}) \Big| \pi\right],
\end{align*}
where the expectation is taken with respect to the state sequence $s_1, s_2, \ldots$ achieved on taking actions according to $\pi$ starting from $s_0$. One can compute an empirical estimate of the feature expectations of the expert's policy with the help of the trajectories $\{Tr^{(1)}, Tr^{(2)}, \ldots, Tr^{(m)}\}$, using
\begin{align}	\label{eqn:est-feat-exp}
\hat{\mu}_E = \frac{1}{m}\sum_{i=1}^m \sum_{t=0}^{L-1} \gamma^t \boldsymbol{\phi}\big(s_t^{(i)}, a_t^{(i)}, s_{t+1}^{(i)}\big).
\end{align}
Given a weight vector $\boldsymbol{w}$, one can compute the optimal policy $\pi_{\boldsymbol{w}}$ for the corresponding reward function $\widehat{\mathcal{R}}_{\boldsymbol{w}}$, and estimate its feature expectations $\hat{\mu}(\pi_{\boldsymbol{w}})$ in a way similar to \eqref{eqn:est-feat-exp}. IRL compares this $\hat{\mu}(\pi_{\boldsymbol{w}})$ with expert's feature expectations $\hat{\mu}_E$ to learn best fitting weight vectors $\boldsymbol{w}$. Instead of a single weight vector, the IRL algorithm by \citet{abbeel2004apprenticeship} learns a set of possible weight vectors, and they ask the agent designer to pick the most appropriate weight vector among these by inspecting their corresponding policies.

\subsection{Contextual Bandits}
Following \citet{11}, the contextual bandit problem is defined as follows.
At each time $t \in \{0,1,\ldots,(T-1)\}$, the player is presented with a {\em context vector} $c(t) \in \mathbb{R}^d$ and must choose an arm $k \in [K] = \{ 1,2,\ldots,K\}$.
%
Let ${\bf r} = (r_{1}(t),\ldots, r_{K}(t))$ denote  a reward vector, where $r_k(t)$ is the reward at time $t$ associated with the arm $k \in [K]$.
%
We assume that the expected reward is a linear function of the context, i.e. $\mathbb{E}[r_k(t)|c(t)] = \mu_k^T c(t)$, where $\mu_k$ is an unknown weight vector (to be learned from the data) associated with the arm $k$.

The purpose of a contextual bandit algorithm $A$ is to minimize the cumulative regret. Let $H:C\rightarrow [K]$ where $C$ is the set of possible contexts and $c(t)$ is the context at time $t$, $h_t \in H$ a hypothesis computed by the algorithm $A$ at time $t$ and $h^{*}_{t}=\underset{h_t \in H}{\operatorname{argmax}}\ r_{h_{t}(c(t))}(t)$ the optimal hypothesis at the same round. The cumulative regret is: $R(T) = \sum ^{T}_{t=1} {r_{h^{*}_{t}(c(t))}(t)- r_{h_t(c(t))}(t)}$.

One widely used way to solve the contextual bandit problem is the Contextual Thompson Sampling algorithm (CTS) \cite{AgrawalG13} given as Algorithm \ref{alg:CTS}.
In CTS, the reward $r_{k}(t)$ for choosing arm $k$ at time $t$ follows a parametric likelihood function $Pr(r(t)|\tilde{\mu})$.  
Following \citet{AgrawalG13}, the posterior distribution at time $t + 1$, $Pr(\tilde{\mu}|r(t)) \propto Pr(r(t)|\tilde{\mu}) Pr(\tilde{\mu})$ is given by a multivariate Gaussian distribution $\mathcal{N}(\hat{\mu_k}(t+1)$, $v^2 B_k(t + 1)^{-1})$, where
$B_k(t)= I_d + \sum^{t-1}_{\tau=1} c(\tau) c(\tau)^\top$, $d$ is the size of the context vectors $c$, $v= R \sqrt{\frac{24}{z} d \cdot ln(\frac{1}{\gamma})}$ and we have $R>0$,  $z \in [0,1]$, $\gamma \in [0,1]$ constants, and $\hat{\mu}(t)=B_k(t)^{-1} (\sum^{t-1}_{\tau=1} c(\tau) r_k(\tau))$.

\begin{algorithm}
   \caption{Contextual Thompson Sampling Algorithm}
\label{alg:CTS}
\begin{algorithmic}[1]
 \STATE {\bfseries }\textbf{Initialize:}  $B_k = I_d$, $ \hat{\mu}_k = 0_d, f_k = 0_d$ for $k \in [K]$.
 \STATE {\bfseries }\textbf{Foreach} $t = 0, 1, 2, \ldots ,(T-1)$ \textbf{do}
 \STATE {\bfseries }\quad Sample $\tilde{\mu_{k}}(t)$ from $N(\hat{\mu}_k, v^2 B_k^{-1})$.
 \STATE {\bfseries }\quad Play arm $k_t= \underset{k\in [K]}{argmax}\ c(t)^\top \tilde{\mu_{k}}(t) $
  \STATE {\bfseries }\quad Observe $r_{k_t}(t)$ 
 \STATE \quad $B_{k_t} = B_{k_t}+ c(t)c(t)^{T} $, $f_{k_t} = f_{k_t} + c(t)r_{k_t}(t)$, \\\quad$\hat{\mu}_{k_t} = B_{k_t}^{-1} f_{k_t}$
 \STATE {\bfseries }\textbf{End}
   \end{algorithmic}
\end{algorithm}

Every step $t$ consists of generating a $d$-dimensional sample $\tilde{\mu_k}(t)$ from $\mathcal{N}(\hat{\mu_k}(t)$, $ v^2B_k(t)^{-1})$ for each arm.  We then decide which arm $k$ to pull by solving for $\operatorname{argmax}_{k\in [K]} c(t)^\top \tilde{\mu_k}(t)$.  This means that at each time step we are selecting the arm that we expect to maximize the observed reward given a sample of our current beliefs over the distribution of rewards, $c(t)^\top \tilde{\mu_k}(t)$.  We then observe the actual reward of pulling arm $k$, $r_k(t)$ and update our beliefs.

\subsection{Problem Setting}

In our setting, the agent is in multi-objective Markov decision processes (MOMDPs), instead of the usual scalar reward function $R(s,a,s')$, a reward vector $\vec{R}(s,a,s')$ is present. The vector $\vec{R}(s,a,s')$ consists of $l$ dimensions or components representing the different objectives, i.e.,  $\vec{R}(s,a,s') =(R_1(s,a,s'),\ldots,R_{l}(s,a,s'))$. However, not all components of the reward vector are observed in our setting. There is an objective $v \in [l]$ that is hidden, and the agent is only allowed to observe expert demonstrations to learn this objective. These demonstrations are given in the form of trajectories $\{Tr^{(1)}, Tr^{(2)}, \ldots, Tr^{(m)}\}$. To summarize, for some objectives, the agent has rewards observed from interaction with the environment, and for some objectives the agent has only expert demonstrations. The aim is still the same as single objective reinforcement learning, which is trying to maximize $\sum_{t=0}^{\infty} \gamma^t R_{i}(s_t,a_t, s_{t+1})$ for each $i \in [l]$.
 

\section{Approach}

\subsection{Domain}
\begin{figure}[t]
  \centering
  \includegraphics[width=0.9\columnwidth]{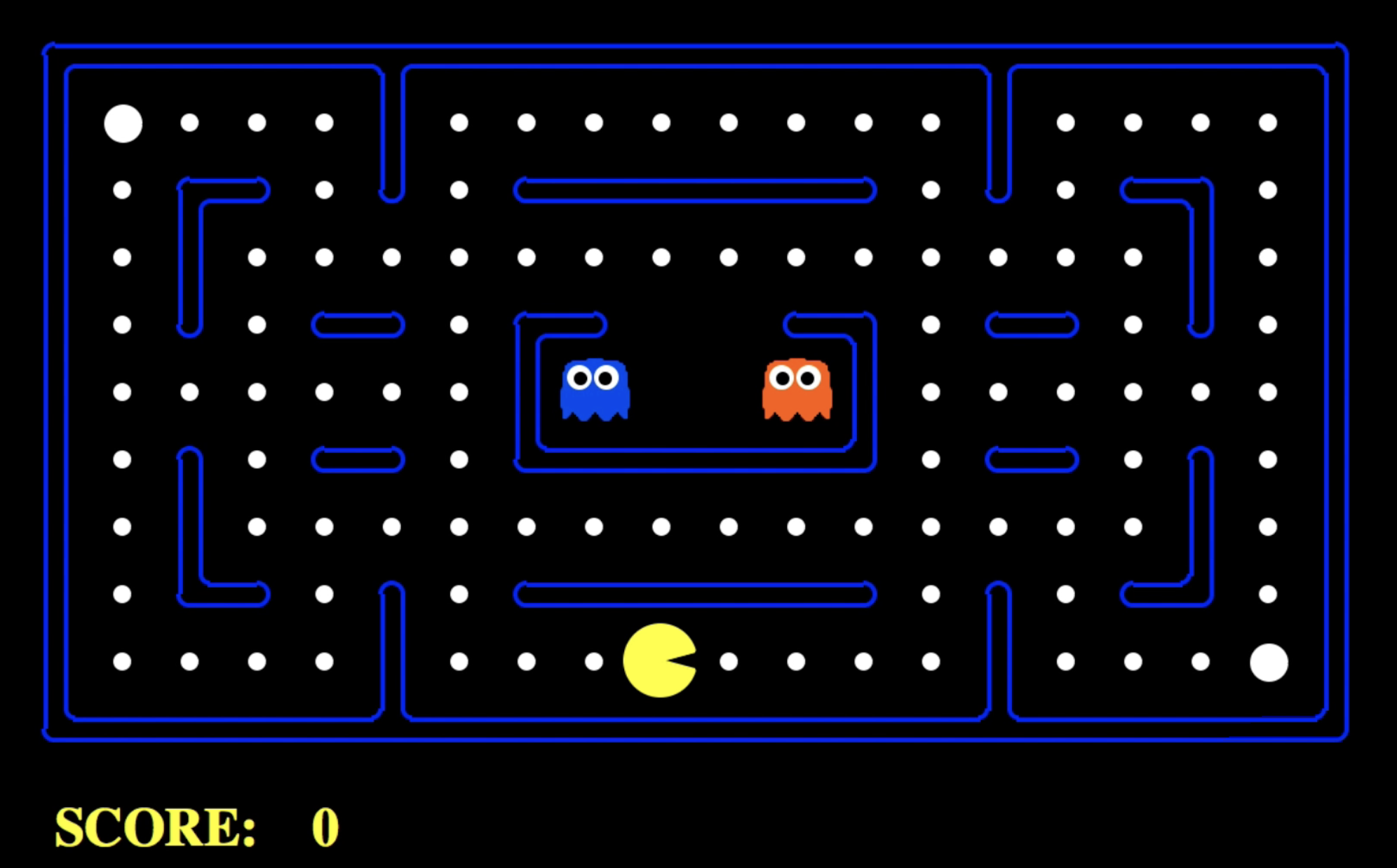}
  \caption{Layout of Pac-Man}
  \label{fig:pacman}
\end{figure}

We demonstrate the applicability of our approach using the classic game of Pac-Man. The layout of Pac-Man we use for this is given in Figure~\ref{fig:pacman}, and the following are the rules used for the environment (adopted from Berkeley AI Pac-Man\footnote{ \url{http://ai.berkeley.edu/project_overview.html}}). The goal of the agent (which controls Pac-Man's motion) is to eat all the dots in the maze, known as Pac-Dots, as soon as possible while simultaneously avoiding collision with ghosts. On eating a Pac-Dot, the agent obtains a reward of $+10$. And on successfully winning the game (which happens on eating all the Pac-Dots), the agent obtains a reward of $+500$. In the meantime, the ghosts in the game roam the maze trying to kill Pac-Man. On collision with a ghost, Pac-Man loses the game and gets a reward of $-500$. The game also has two special dots called capsules or Power Pellets in the corners of the maze, which on consumption, give Pac-Man the temporary ability of ``eating'' ghosts. During this phase, the ghosts are in a ``scared'' state for 40 frames and move at half their speed. On eating a ghost, the agent gets a reward of $+200$, the ghost returns to the center box and returns to its normal ``unscared'' state.
	Finally, there is a constant time-penalty of $-1$ for every step taken.

For the sake of demonstration of our approach, we define \textit{not eating ghosts} as the desirable constraint in the game of Pac-Man. However, please recall that this constraint is not given explicitly to the agent, but only through examples. To play optimally in the original game one should eat ghosts to earn bonus points, but doing so is being demonstrated as undesirable. Hence, the agent has to combine the goal of collecting the most points while not eating ghosts if possible.

\subsection{Overall Approach}	\label{subsec:overall}

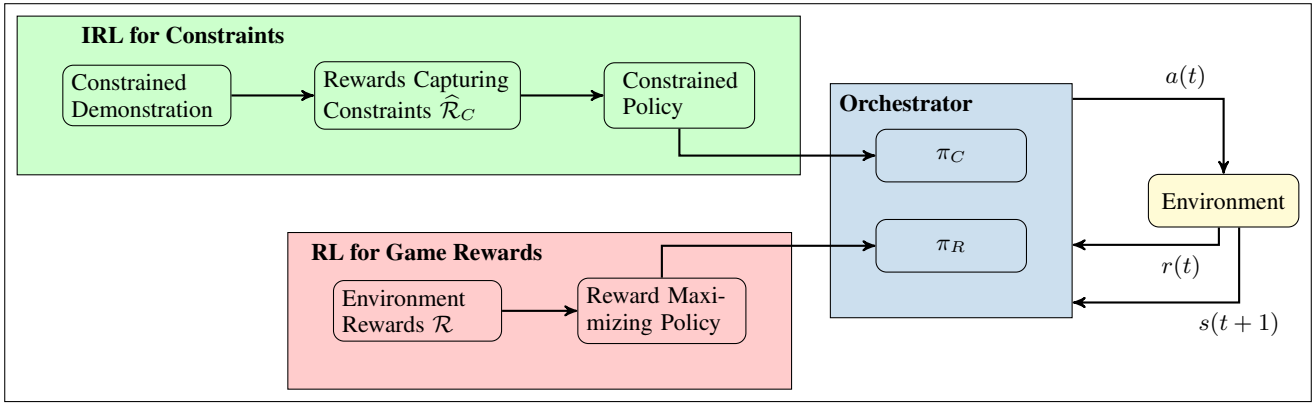
\begin{figure*}[ht]
  \centering
  \begin{tikzpicture}[show background rectangle, font=\footnotesize]

    \node[module, text width=2cm] (I1) {Constrained Demonstration};
    \node[module, right=of I1,xshift=1mm, text width=2.5cm] (I2) {Rewards Capturing Constraints $\widehat{\mathcal{R}}_C$};
    \node[module, right=of I2,xshift=1mm, text width=1.5cm] (I3) {Constrained Policy};
    
    \node[module, below=of I2, text width=2cm, yshift=-10mm] (R1) {Environment Rewards $\mathcal{R}$};
    \node[module, right=of R1, text width=2cm] (R2) {Reward Maximizing Policy};
    
	\node[module, right=of I3,xshift=6mm, yshift=-8mm, inner sep=2mm] (pc) {$\pi_C$};
	\node[module, right=of I3,xshift=6mm, yshift=-20mm, inner sep=2mm] (pr) {$\pi_R$};
	
 
    \begin{pgfonlayer}{background}
    	\node[fit=(I1) (I2) (I3), draw, fill=green!20, 
        inner sep=6mm, label={[xshift=-30mm,yshift=-5mm] \textbf{IRL for Constraints}}] (IRL) {};
	
	    \node[fit=(R1) (R2), draw, fill=red!20, 
         inner sep=6mm, label={[xshift=-15mm,yshift=-5mm] \textbf{RL for Game Rewards}}] (RL) {};

        \node[fit=(pc) (pr), draw, fill=blue!20, 
         inner sep=6mm, label={[xshift=-6mm,yshift=-5mm] \textbf{Orchestrator}}] (orc) {};

    \end{pgfonlayer}
    
    \node[module, right=of orc, fill=yellow!20] (env) {Environment};

    \draw[beats] (I1) to (I2);
    \draw[beats] (I2) -- node [above,yshift=1mm] {} (I3);
    \draw[beats] (I3) |- node [above,xshift=20mm] {} (pc);
    
    \draw[beats] (R1) to (R2);
    \draw[beats] (R2) |- node [below,xshift=18mm] {} (pr);
    
    \draw[beats] (orc.40) -| node [above, xshift=-5mm] {$a(t)$} (env.north);
    \draw[beats] (env.300) |- node [below] {$s(t+1)$} (orc.320);
        \draw[beats] (env.260) |- node [below, xshift=-5mm] {$r(t)$} (orc.340);


\end{tikzpicture}

  \caption{Overview of our system.  At each time step the Orchestrator selects between two policies, $\pi_C$ and $\pi_R$ depending on the observations from the Environment.  The two policies are learned before engaging with the environment. $\pi_C$ is obtained using IRL on the demonstrations to learn a reward function that captures the particular constraints demonstrated. The second, $\pi_R$ is obtained by the agent through RL on the environment directly. }
  \label{fig:pipeline}
  \end{figure*}

The overall approach we follow is depicted by Figure~\ref{fig:pipeline}. It has three main components. The first is the inverse reinforcement learning component to learn the desirable constraints (depicted in green in Figure~\ref{fig:pipeline}). We apply inverse reinforcement learning to the demonstrations depicting desirable behavior, to learn the underlying constraint rewards being optimized by the demonstrations. We then apply reinforcement learning on these learned rewards to learn a strongly constraint satisfying policy $\pi_C$.

Next, we augment this with a pure reinforcement learning component (depicted in red in Figure~\ref{fig:pipeline}). For this, we directly apply reinforcement learning to the original environment rewards (like Pac-Man's unmodified game) to learn a domain reward maximizing policy $\pi_R$. Just to recall, the reason we have this second component is that the inverse reinforcement learning component may not be able to pick up the original environment rewards very well since the demonstrations were intended mainly to depict desirable behavior. Further, since these demonstrations are given by humans, they are prone to error, amplifying this issue. Hence, the constraint obeying policy $\pi_C$ is likely to exhibit strong constraint satisfying behavior, but may not be optimal in terms of maximizing environment rewards. Augmenting with the reward maximizing policy $\pi_R$ will help the system in this regard.

So now, we have two policies, the constraint-obeying policy $\pi_C$ and the reward-maximizing policy $\pi_R$. To combine these two, we use the third component, the orchestrator (depicted in blue in Figure~\ref{fig:pipeline}). This is a contextual bandit algorithm that orchestrates the two policies, picking one of them to play at each point of time. The context is the state of the environment (state of the Pac-Man game); the bandit decides which arm (policy) to play at the corresponding point of time.

\subsection{Alternative Approaches}
Observe that in our approach, we combine or ``aggregate'' the two objectives (environment rewards and desired constraints) at the policy stage. Alternative approaches to doing this are combining the two objectives at the reward stage or the demonstrations stage itself:
\begin{itemize}
	\item \textbf{Aggregation at reward phase.} As before, we can perform inverse reinforcement learning to learn the underlying rewards capturing the desired constraints. Now, instead of learning a policy for each of the two reward functions (environment rewards and constraint rewards) followed by aggregating them, we could just combine the reward functions themselves. And then, we could learn a policy on this ``aggregated'' rewards to perform well on both the objectives, environment reward and favorable constraints. (This captures the intuitive idea of ``incorporating the constraints into the environment rewards'' if we were explicitly given the penalty of violating constraints).
   
    \item \textbf{Aggregation at data phase.} Moving another step backward, we could aggregate the two objectives of play at the data phase. This could be performed as follows. We perform pure reinforcement learning as in the original approach given in Figure~\ref{fig:pipeline} (depicted in red). Once we have our reward maximizing policy $\pi_R$, we use it to generate numerous reward-maximizing demonstrations. Then, we combine these environment reward trajectories with the original constrained demonstrations, aggregating the two objectives in the process. And once we have the combined data, we can perform inverse reinforcement learning to learn the appropriate rewards, followed by reinforcement learning to learn the corresponding policy.
\end{itemize}

Aggregating at the policy phase is where we go all the way to the end of the pipeline learning a policy for each of the objectives,  followed by aggregating them. This is the approach we follow as mentioned in Section~\ref{subsec:overall}. Note that, we have a parameter $\lambda$ (as described in more detail in Section \ref{sec:orch}) that trades off environmental rewards and rewards capturing constraints. A similar parameter can be used by the reward aggregation and data aggregation approaches, to decide how to weigh the two objectives while performing the corresponding aggregation.

The question now is, ``which of these aggregation procedures is the most useful?''. The reason we use aggregation at the policy stage is to gain \textit{interpretability}. Using an orchestrator to pick a policy at each point of time helps us identify which policy is being played at each point of time and also the reason for which it is being chosen (in the case of an interpretable orchestrator, which it is in our case). More details on this are mentioned in Section~\ref{sec:expts}.

\section{Concretizing Our Approach}

Here we describe the exact algorithms we use for each of the components of our approach.

\subsection{Details of the Pure RL}

For the reinforcement learning component, we use Q-learning with linear function approximation as described in Section~\ref{subsec:background-rl}. For Pac-Man, some of the features we use for an $\langle s,a\rangle$ pair (for the $\boldsymbol{\psi}(s,a)$ function) are: ``whether food will be eaten'', ``distance of the next closest food'', ``whether a scared (unscared) ghost collision is possible'' and ``distance of the closest scared (unscared) ghost''.

For the layout of Pac-Man we use (shown in Figure~\ref{fig:pacman}), an upper bound on the maximum score achievable in the game is $2170$. This is because there are $97$ Pac-Dots, each ghost can be eaten at most twice (because of two capsules in the layout), Pac-Man can win the game only once and it would require more than $100$ steps in the environment. On playing a total of $100$ games, our reinforcement learning algorithm (the reward maximizing policy $\pi_R$) achieves an average game score of $1675.86$, and the maximum score achieved is $2144$. We mention this here, so that the results in Section~\ref{sec:expts} can be seen in appropriate light.

\subsection{Details of the IRL}
For inverse reinforcement learning, we use the linear IRL algorithm as described in Section~\ref{subsec:background-irl}. For Pac-Man, observe that the original reward function $\mathcal{R}(s,a,s')$ depends only on the following factors: ``number of Pac-Dots eating in this step $(s,a,s')$'', ``whether Pac-Man has won in this step'', ``number of ghosts eaten in this step'' and ``whether Pac-Man has lost in this step''. For our IRL algorithm, we use exactly these as the features $\boldsymbol{\phi}(s,a,s')$. As a sanity check, when IRL is run on environment reward optimal trajectories (generated from our policy $\pi_R$), we recover something very similar to the original reward function $\mathcal{R}$. In particular, the weights of the reward features learned is given by
$$\frac{1}{1000}[+2.44, +138.80, +282.49, -949.17],$$
which when scaled is almost equivalent to the true weights $[+10, +500, +200, -500]$ in terms of their optimal policies. The number of trajectories used for this is $100$.

Ideally, we would prefer to have the constrained demonstrations given to us by humans. But for our domain of Pac-Man, we generate them synthetically as follows. We learn a policy $\pi_C^\star$ by training it on the game with the original reward function $\mathcal{R}$ augmented with a very high negative reward ($-1000$) for eating ghosts. This causes $\pi_C^\star$ to play well in the game while avoiding eating ghosts as much as possible.\footnote{We do this only for generating demonstrations. In real domains, we would not have access to the exact constraints that we want to be satisfied, and hence a policy like $\pi_C^\star$ cannot be learned; learning from human demonstrations would then be essential.} Now, to emulate erroneous human behavior, we use $\pi_C^\star$ with an error probability of $3\%$. That is, at every time step, with $3\%$ probability we pick a completely random action, and otherwise follow $\pi_C^\star$. This gives us our constrained demonstrations, on which we perform inverse reinforcement learning to learn the rewards capturing the constraints. The weights of the reward function learned is given by
$$\frac{1}{1000} [+2.84, +55.07, -970.59, -234.34],$$
and it is evident that it has learned that eating ghosts strongly violates the favorable constraints. The number of demonstrations used for this is $100$. We scale these weights to have a similar $L_1$ norm as the original reward weights $[+10, +500, +200, -500]$, and denote the corresponding reward function by $\widehat{\mathcal{R}}_C$.

Finally, running reinforcement learning on these rewards $\widehat{\mathcal{R}}_C$, gives us our constraint policy $\pi_C$. On playing a total of $100$ games, $\pi_C$ achieves an average game score of $1268.52$ and eats just $0.03$ ghosts on an average. Note that, when eating ghosts is prohibited in the domain, an upper bound on the maximum score achievable is $1370$.

\subsection{Orchestration with Contextual Bandits}
\label{sec:orch}

We use contextual bandits to pick one of the policies ($\pi_R$ and $\pi_C$) to play at each point of time. These two policies act as the two arms of the bandit, and we use a modified CTS algorithm to train the bandit. The context of the bandit is given by features of the current state (for which we want to decide which policy to choose), i.e., $c(t) = \Upsilon(s_t) \in \mathbb{R}^d$. For the game of Pac-Man, the features of the state we use for context $c(t)$ are: (i) A constant $1$ to represent the bias term, and (ii) The distance of Pac-Man from the closest scared ghost in $s_t$. One could use a more sophistical context with many more features, but we use this restricted context to demonstrate a very interesting behavior (shown in Section~\ref{sec:expts}).

The exact algorithm used to train the orchestrator is given in Algorithm~\ref{alg:orchestrator}. Apart from the fact that arms are policies (instead of atomic actions), the main difference from the CTS algorithm is the way rewards are fed into the bandit. For simplicity, we call the constraint policy $\pi_C$ as arm $0$ and the reward policy $\pi_R$ as arm $1$. We now go over Algorithm~\ref{alg:orchestrator}. First, all the parameters are initialized as in the CTS algorithm (Line 1). For each time-step in the training phase (Line 3), we do the following. Pick an arm $k_t$ according to the Thompson Sampling algorithm and the context $\Upsilon(s_t)$ (Lines 4 and 5). Play the action according to the chosen policy $\pi_{k_t}$ (Line 6). This takes us to the next state $s_{t+1}$. We also observe two rewards (Line 7): (i) the original reward in environment, $r^R_{a_t}(t) = \mathcal{R}(s_t, a_t, s_{t+1})$ and (ii) the constraint rewards according to the rewards learnt by inverse reinforcement learning, i.e., $r^C_{a_t}(t) = \widehat{\mathcal{R}}_C(s_t, a_t, s_{t+1})$. $r^C_{a_t}(t)$ can intuitively be seen as the predicted reward (or penalty) for any constraint satisfaction (or violation) in this step.

\begin{algorithm}[h!]
  \caption{Orchestrator Based Algorithm}
\label{alg:orchestrator}
\begin{algorithmic}[1]
 \STATE \textbf{Initialize:}  $B_k = I_d$, $ \hat{\mu}_k = 0_d, f_k = 0_d$ for $k \in \{0,1\}$.
 \STATE \textbf{Observe} start state $s_0$.
 \STATE \textbf{Foreach} $t = 0, 1, 2, . . . ,(T-1)$ \textbf{do}
 \STATE \quad Sample $\tilde{\mu_{k}}(t)$ from $N(\hat{\mu}_k, v^2 B_k^{-1})$.
 \STATE \quad Pick arm $k_t = \underset{k \in \{0,1\}}{argmax}\ \Upsilon(s_t)^\top \tilde{\mu_{k}}(t)$.
 \STATE \quad Play corresponding action $a_t = \pi_{k_t}(s_t)$.
 \STATE \quad Observe rewards $r^C_{a_t}(t)$ and $r^R_{a_t}(t)$, and the next state $s_{t+1}$.
 \STATE \quad Define $r_{k_t}(t) = \lambda\left(r^C_{a_t}(t) + \gamma V^C(s_{t+1})\right)$ \\\qquad\quad\qquad\qquad $+ (1-\lambda)\left(r^R_{a_t}(t) + \gamma V^R(s_{t+1})\right)$
 \STATE \quad Update $B_{k_t} = B_{k_t} + \Upsilon(s_t)\Upsilon(s_t)^\top$, $f_{k_t} = f_{k_t} + \Upsilon(s_t) r_{k_t}(t)$, $\hat{\mu}_{k_t} = B_{k_t}^{-1} f_{k_t}$
 \STATE \textbf{End}
   \end{algorithmic}
\end{algorithm}

To train the contextual bandit to choose arms that perform well on both metrics (environment rewards and constraints), we feed it a reward that is a linear combination of $r^R_{a_t}(t)$ and $r^C_{a_t}(t)$ (Line 8). Another important point to note is that $r^R_{a_t}(t)$ and $r^C_{a_t}(t)$ are immediate rewards achieved on taking action $a_t$ from $s_t$, they do not capture long term effects of this action. In particular, it is important to also look at the ``value'' of the next state $s_{t+1}$ reached, since we are in the sequential decision making setting. Precisely for this reason, we also incorporate the value-function of the next state $s_{t+1}$ according to both the reward maximizing component and constraint component (which encapsulate the long-term rewards and constraint satisfaction possible from $s_{t+1}$). This gives exactly Line 8, where $V^C$ is the value-function according the constraint policy $\pi_C$, and $V^R$ is the value-function according to the reward maximizing policy $\pi_R$. In this equation, $\lambda$ is a hyperparameter chosen by a user to decide how much to trade off environment rewards for constraint satisfaction. For example, when $\lambda$ is set to $0$, the orchestrator would always play the reward policy $\pi_R$, while for $\lambda = 1$, the orchestrator would always play the constraint policy $\pi_C$. For any value of $\lambda$ in-between, the orchestrator is expected to pick policies at each point of time that would perform well on both metrics (weighed according to $\lambda$). Finally, for the desired reward $r_{k_t}(t)$ and the context $\Upsilon(s_t)$, the parameters of the bandit are updated according to the CTS algorithm (Line 9).

\section{Evaluation and Test}	\label{sec:expts}

\begin{figure}[t]
  \centering
  \includegraphics[width=0.9\columnwidth]{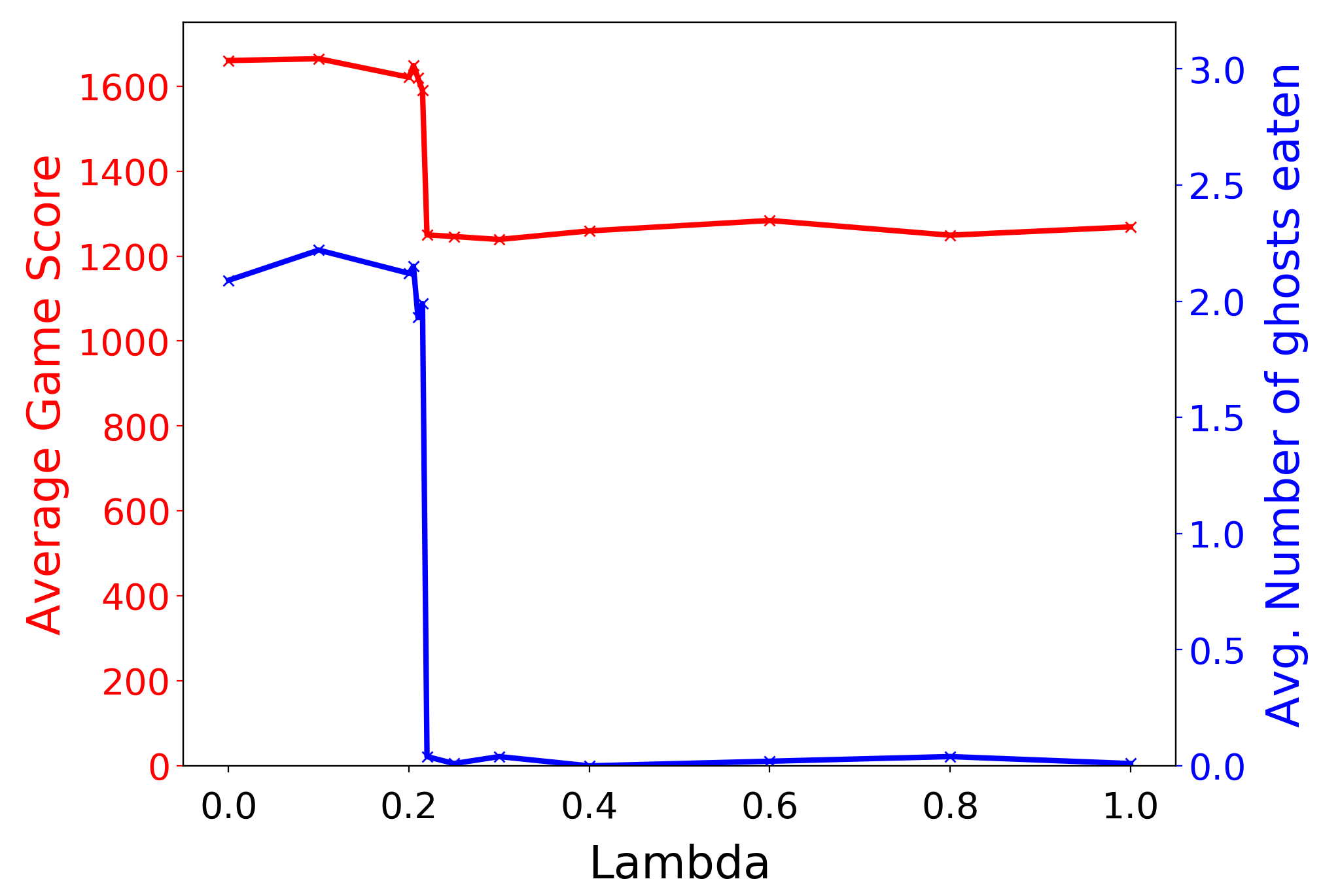}
  \caption{Both performance metrics as $\lambda$ is varied. The red curve depicts the average game score achieved, and the blue curve depicts the average number of ghosts eaten.}
  \label{fig:both-metrics}
\end{figure}

We test our approach on the Pac-Man domain given in Figure~\ref{fig:pacman}, and measure its performance on two metrics, (i) the total score achieved in the game (the environment rewards) and (ii) the number of ghosts eaten (the constraint violation). We also vary $\lambda$, and observe how these metrics are traded off against each other. For each value of $\lambda$, the orchestrator is trained for $100$ games. The results are shown in Figure~\ref{fig:both-metrics}. Each point in the graph is averaged over $100$ test games.

The graph shows a very interesting pattern. When $\lambda$ is at most than $0.215$, the agent eats a lot of ghosts, but when it is above $0.22$, it eats almost no ghosts. In other words, there is a value $\lambda_o \in [0.215, 0.22]$ which behaves as a tipping point, across which there is drastic change in behavior. Beyond the threshold, the agent learns that eating ghosts is not worth the score it is getting and so it avoids eating as much as possible. On the other hand, when $\lambda$ is smaller than this threshold, it learns the reverse and eats as many ghosts as possible.

\smallskip
\noindent
\textbf{Policy-switching.} As mentioned before, one of the most important property of our approach is interpretability, we know exactly which policy is being played at each point of time. For moderate values of $\lambda$, the orchestrator learns a very interesting policy-switching technique: whenever at least one of the ghosts in the domain is scared, it plays $\pi_C$, but if no ghosts are scared, it plays $\pi_R$. In other words, it starts off the game by playing $\pi_R$ until a capsule is eaten. As soon as the first capsule is eaten, it switches to $\pi_C$ and plays it till the scared timer runs off. Then it switches back to $\pi_R$ until another capsule is eaten, and so on.\footnote{A video of our agent demonstrating this behavior is uploaded in the Supplementary Material. The agent playing the game in this video was trained with $\lambda = 0.4$.} It has learned a very intuitive behavior: when there is no scared ghost in the domain, there is no possibility of violating constraints, and hence the agent is as greedy as possible (i.e., play $\pi_R$), but when there are scared ghosts, better to be safe (i.e., play $\pi_C$).

\section{Discussion}

In this paper, we have considered the problem of autonomous agents learning policies that are constrained by implicitly-specified norms and values while still optimizing their policies with respect to environmental rewards.  We have taken an approach that combines IRL to determine constraint-satisfying policies from demonstrations, RL to determine reward-maximizing policies, and a contextual bandit to orchestrate between these policies in a transparent way.  This proposed architecture and approach for the problem is novel. It also requires a novel technical contribution in the contextual bandit algorithm because the arms are policies rather than atomic actions, thereby requiring rewards to account for sequential decision making.  We have demonstrated the algorithm on the Pac-Man video game and found it to perform interesting switching behavior among policies.

We feel that the contribution herein is only the starting point for research in this direction.  We have identified several avenues for future research, especially with regards to IRL.  We can pursue deep IRL to learn constraints without hand-crafted features, develop an IRL that is robust to noise in the demonstrations, and research IRL algorithms to learn from just one or two demonstrations (perhaps in concert with knowledge and reasoning). In real-world settings, demonstrations will likely be given by different users with different versions of abiding behavior; we would like to exploit the partition of the set of traces by user to improve the policy or policies learned via IRL.  Additionally, the current orchestrator selects a single policy at each time, but more sophisticated policy aggregation techniques for combining or mixing policies is possible. Lastly, it would be interesting to investigate whether the policy aggregation rule ($\lambda$ in the current proposal) can be learned from demonstrations.

\subsubsection{Acknowledgments}
We would like to thank Gerald Tesauro and Aleksandra Mojsilovic for their helpful feedback and comments on this project.

\bibliographystyle{aaai}

\end{document}